\documentclass[twoside,leqno,twocolumn]{article}
\usepackage{ltexpprt}
\usepackage{booktabs} 
\usepackage{kotex}

\usepackage{graphicx}
\usepackage{graphics}
\usepackage{booktabs}
\usepackage{multirow}
\usepackage{tabulary}

\usepackage{dsfont}

\begin{document}

\title{\Large Predicting Multiple Demographic Attributes with Task Specific Embedding Transformation and Attention Network}
\author{Raehyun Kim\thanks{These authors contributed equally to this work. Department of Computer Science and Engineering, Korea University. \{raehyun, hyunjae-kim\}@korea.ac.kr} \\
\and
Hyunjae Kim\footnotemark[1] \\
\and
Janghyuk Lee\thanks{Business School, Korea University. janglee@korea.ac.kr}
\and
Jaewoo Kang\thanks{Corresponding author. Department of Computer Science and Engineering, Korea University. kangj@korea.ac.kr} 
}
\date{}

\maketitle


\fancyfoot[R]{\scriptsize{Copyright \textcopyright\ 2019 by SIAM\\
Unauthorized reproduction of this article is prohibited}}





\begin{abstract} \small\baselineskip=9pt Most companies utilize demographic information to develop their strategy in a market. However, such information is not available to most retail companies. Several studies have been conducted to predict the demographic attributes of users from their transaction histories, but they have some limitations. First, they focused on parameter sharing to predict all attributes but capturing task-specific features is also important in multi-task learning.
Second, they assumed that all transactions are equally important in predicting demographic attributes. 
However, some transactions are more useful than others for predicting a certain attribute.
Furthermore,  decision making process of models cannot be interpreted as they work in a black-box manner. 
To address the limitations, we propose an Embedding Transformation Network with Attention (ETNA) model which shares representations at the bottom of the model structure and transforms them to task-specific representations using a simple linear transformation method.
In addition, we can obtain more informative transactions for predicting certain attributes using the attention mechanism.
The experimental results show that our model outperforms the previous models on all tasks.
In our qualitative analysis, we show the visualization of attention weights, which provides business managers with some useful insights.
\end{abstract}

\section{Introduction}
\label{section:intro}

Customers' demographic information is a valuable asset for many companies as it is useful for making strategic decisions. When developing their marketing strategy, most companies use demographic segmentation to decide which market to target \cite{lin2002segmenting}. For example, Apple's iPhone  targets mainly a relatively young customers (age) with enough purchasing power (income). Nowadays, as the importance of successful recommender systems grows, numerous methods have used demographic information to improve the quality of the systems  and solve the cold start problem \cite{gupta2015performance, safoury2013exploiting}.

The methods mentioned above can be successfully used only when there is sufficient demographic information. However, it is difficult for most companies to collect demographic information of customers. Due to a series of data breaches, customers are reluctant to give their sensitive information to companies. As Wang \textit{et al.} \cite{wang2016your} pointed out, in a real world data, only partial demographic attributes are known for a great number of users and some users have no attributes at all.

Demographic information is used to determine purchase propensity in aforementioned methods. If demographic information contains purchase propensity data, purchasing history can be used to predict demographic information as well. For example, Wang \textit{et al.} \cite{wang2016your}  focused on predicting the demographic attributes of customers using their purchasing histories. Resheff \textit{et al.} \cite{resheff2017fusing} used a similar approach but employed a model with a deeper structure. 

Although they raised interesting questions and made some meaningful achievements, their works have some limitations. First, although we consider demographic attributes prediction as a multi-task problem, the previous works focused on learning shared user representation from user transactions. By sharing representation, models could learn patterns and trends of  users. For example, users who purchase products popular in the older age group are more likely to be married. However, in multi-task learning, it is also important that the model is designed to learn more task-specific features. No existing works have achieved a balance between learning general features and task-specific features. Second, existing works assume that all transactions are equally important for demographic prediction, which is not true in most settings. Intuitively, purchasing multiple cosmetic items can be a clue that the user is female. However, daily necessities do not provide much information for inferring a user's gender. Some transactions contribute more to predict demographic attribute than others and the importance varies for each task. Additionally, although the existing models improved the prediction accuracy, none of them have adopted an interpretable model structure. In real world business, understanding how a model predicts answers provides us with useful insights on consumer behavior.

To address the limitations mentioned above, we propose Embedding Transformation Network with Attention (ETNA) to predict demographic attributes from transaction data.  In ETNA, we  share embeddings at the bottom of the model structure and convert them to task-specific embeddings using simple linear transformation. This transformation selectively captures appropriate features for each task from shared information. To focus more on informative transactions than irrelevant ones, we apply attention mechanism to each task. The experimental results demonstrate that our model is far more accurate than existing models. Furthermore, by analyzing and visualizing attention mechanism, we can identify which type of transaction contributed more to predict an attribute. Moreover, we release a new benchmark dataset for demographic prediction in retail business scenario. 

To summarize, our work makes following contributions:
\begin{enumerate}
\item We propose Embedding Transformation Network with Attention (ETNA) that learns task-specific features from shared information and automatically discriminating transactions that are more informative for predicting an attribute.

\item The experimental results show that our model is far more accurate than existing models. Using our attention network's scoring mechanism, we conducted qualitative analysis and obtained results that provide insight into consumer behavior.

\item We developed and release a new benchmark dataset for demographic prediction in retail business scenario which could be used for future research.
\end{enumerate}

The remainder of the paper is organized as follows. After discussing related works in Section \ref{section:related work}, we provide the task description in Section \ref{section: task description}. In Section \ref{section: our approach}, we describe our proposed ETNA model in detail. The experiment results and qualitative analysis are presented in Section \ref{section:exp}. Finally, we  conclude our paper in Section \ref{section:conclustion}. 

\section{Related Work}
\label{section:related work}

\subsection{Demographic prediction}
\label{section:demo pred}

Numerous approaches for predicting the demographic attributes of a user from various types of data have been proposed. Hu \textit{et al.} \cite{hu2007demographic} have used the browsing history of users. They found that search behavior can vary depending on gender and age, and they achieved meaningful prediction accuracy. With the advent of the big data era, several works have used social network and mobile phone data in demographic prediction tasks \cite{bi2013inferring, culotta2015predicting}. Social network data and search queries are used to infer demographic attributes.  Also, location and mobile application usage data have also been used to predict demographic attributes \cite{malmi2016you, zhong2015you}. 

Some works have used purchasing history to predict demographic attributes. Wang \textit{et al.} \cite{wang2016your} used a large scale dataset from a Chinese retailer, which was originally designed for a recommendation task. 
The dataset is publicly available but has no metadata (e.g. item descriptions, categories) that is needed to interpret a model. Resheff \textit{et al.} \cite{resheff2017fusing} also used the purchasing history of users, which is not available to the public. To the best of our knowledge, there is no publicly accessible dataset that contains both metadata and user demographic information. To understand how our model predicts demographic attributes for real world business, we used transaction data with its metadata. We make our transaction data  freely available for public use and future research on demographic prediction. A detailed description of our dataset is provided in Section 5.

\begin{figure*}[t]
    \begin{center}
    \includegraphics[width=16.5cm,height=5.8cm]{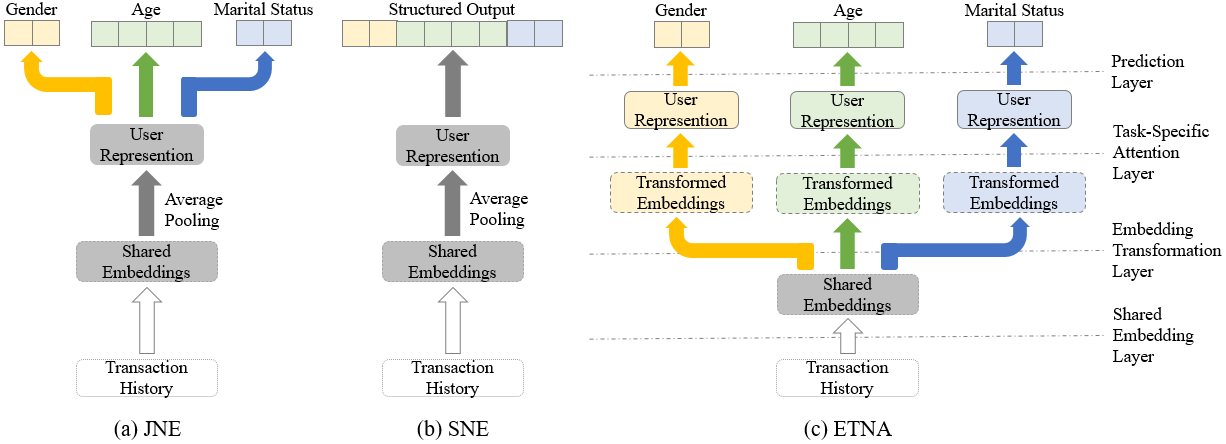}
    \caption{Comparison of model architectures in demographic prediction.}
    \label{fig:model comparison}
    \end{center}
\end{figure*}

\subsection{Multi-Task Learning}
\label{section:multi-task}

Multi-task learning has been widely used in various domains such as natural language processing \cite{luong2015multi}, speech recognition \cite{deng2013new}, computer vision \cite{misra2016cross}, and so on. 

Despite its success in various tasks, there remains some challenges in multi-task learning. In Multi-task learning,  the underlying assumption is that there are features helpful for each task learned from similar tasks. We should decide the type and amount of information to be shared.  There exists a trade-off between sharing general information and capturing task-specific features. When we focus too much on general information by sharing parameters too hard, we lose task-specific signals. Helpful signals from other related tasks may be missed if models are focused on only a single task. One promising approach to solve this problem is to assume that task parameters within a group lie in a low dimensional subspace. Based on this assumption, Kumar et al. \cite{kumar2012learning} argued that each task parameter vector can be obtained by a linear combination of a finite number of underlying basis latent vectors. Inspired by Kumar's work, we use simple transformation to shared representation to obtain task-specific representation.

\subsection{Attention Mechanism}
\label{section:attention mechanism}

Attention mechanism has been used and proven to be successful in various tasks. It was first proposed to capture the most relevant information at each step in a machine translation task \cite{bahdanau2014neural}. Since attention mechanism can selectively focus on more informative data, it is also used in other natural language processing tasks such as question answering \cite{kumar2016ask} and text classification \cite{yang2016hierarchical}. Many other domains such as recommendation and computer vision also adopted attention mechanism  \cite{chen2017attentive, lu2016hierarchical}.

Researchers use attention mechanism not only to improve performance, but also to make more interpretable models. In some domains such as medical diagnosis, designing a model that can be interpreted by a human is as important as achieving high accuracy \cite{zhang2017mdnet}. Furthermore, by analyzing the decision making process of a model, we can obtain useful insights into the domains of interest \cite{luo2018beyond}.

\section{Problem Formalization}
\label{section: task description}

In this section, we formalize a demographic prediction task in retail business scenario. Let $\mathbf{D}$ = [($\mathbf{x}_{1}$, $\mathbf{y}_{1}$),...,($\mathbf{x}_{\mathrm{N}}$, $\mathbf{y}_{\mathrm{N}}$)] be a list of all data samples, where N is the number of samples. As each sample corresponds to an individual user, $\mathbf{x}_n$ and $\mathbf{y}_n$ are the all historical transactions and the demographic attributes of the $n$-th user, respectively. 
$\mathbf{y}_n$ also can be viewed as a list of labels for each task $[y_{n}^{1},...,y_{n}^{\mathrm{M}}]$, where $\mathrm{M}$ is the number of attributes. 
The number of possible classes for the $m$-th attribute is $\mathrm{C}^{m}$.
The transaction history $\mathbf{x}_n$ = [$x_{(n,1)}$,...,$x_{(n,{\mathrm{B}}_n)}$] can be either an ordered and unordered list of transactions depending on data sets, where $\mathrm{B}_n$ is the length of the $n$-th user's history.

In a real world scenario, we might have full or partial, even none information about demographic attributes of users. 
Our goal is to predict all the missing attributes in the dataset.
We follow two types of problem used in \cite{wang2016your}.

\begin{itemize}

\item \textbf{Partial Label Prediction} is for the situation that users gave some part of their demographic attributes (partially observed), so that a company wants to know the remain unknown attributes. Let $\mathbf{X}$ = [$\mathbf{x}_1$,...,$\mathbf{x}_{\mathrm{N}}$] be a set of users' transaction histories and $\mathbf{Y}^{\texttt{obs}}$ = [$\bar{\mathbf{y}}_{1}$,...,$\bar{\mathbf{y}}_{\mathrm{N}}$] be the users' demographic attributes which are partially observed. Given $\mathbf{X}$ with $\mathbf{Y}^{\texttt{obs}}$, the objective is to learn a function to predict the unknown attributes $\mathbf{Y}^{\texttt{unk}}$ = [$\ddot{\mathbf{y}}_{1}$,...,$\ddot{\mathbf{y}}_{N}$].
Note that $\bar{\mathbf{y}}_{n} \cup \ddot{\mathbf{y}}_{n}$ = [$y^{1}_n$,...,$y^{\mathrm{M}}_n$] = $\mathbf{y}_n$.

\item \textbf{New User Prediction} is to predict demographic attributes for new users. Given $\mathbf{X}^{\texttt{obs}}$ with partially/fully observed attributes $\mathbf{Y}^{\texttt{obs}}$ the objective is to learn a function to predict demographic attributes for new users. New users' transaction histories are  $\mathbf{X}^{\texttt{unk}}$ and corresponding labels are $\mathbf{Y}^{\texttt{unk}}$. Note that unlike partial label prediction where $\mathbf{X}$ is used as the input for both training and test sets, $\mathbf{X}$ is splitted into $\mathbf{X}^{\texttt{obs}}$ for the training set and $\mathbf{X}^{\texttt{unk}}$ for the test set, which implies $\mathbf{X}^{\texttt{obs}} \cap \mathbf{X}^{\texttt{unk}} = \emptyset$. 
\end{itemize}

\section{Our Approach}
\label{section: our approach}

In this section, we explain our Embedding Transformation Network with Attention (ETNA) in detail.
First, we learn embeddings of transactions which are shared across all tasks in a Shared Embedding Layer. The embeddings are learned for each item or company of a purchased item depending on the dataset. For our dataset, we learned embeddings for each company.
To obtain task-specific representations, we use an Embedding Transformation Layer that converts a shared vector space into a task-specific vector space for each task using a simple linear transformation.
To encode relationships between transactions and demographic attributes, we opt for an attention mechanism in Task-specific Attention Layer.
Finally, we obtain prediction values for each class from Prediction Layer. The overall model architecture is depicted in Figure \ref{fig:model comparison}-(c). 

Note that given the user's transaction history $\mathbf{x}_n$ = [$x_{(n,1)}$, ...,$x_{(n,{\mathrm{B}}_n)}$], the task is to predict all their unknown attributes ${\ddot{\mathbf{y}}}_n$=[$\ddot{y}^{1}_n$, ...,$\ddot{y}^{\mathrm{M}}_n$]. In partial label prediction problem, the number of attributes to be predicted is less than or equal to $\mathrm{M}$, but to simplify the description we fix the number of unknown attributes to $\mathrm{M}$. We also replace $\ddot{\mathbf{y}}_n$ with $\mathbf{y}_n$ and omit the subscript $n$.

\subsection{Shared Embedding Layer}
\label{section:shared embedding layer}

Given a transaction history $\mathbf{x} = [x_1,...,x_{\mathrm{B}}]$ where each transaction is represented as an index of an item or company, a shared embedding layer maps all transaction to $k$-dimensional vectors. $\mathrm{E} \in \mathcal{R}^{\mathrm{B} \times \mathrm{k}}$ is obtained by this operation.
In multi-task learning, the underlying assumption is that there are features helpful for each task learned from similar tasks. 
Accordingly, in shared embedding layer, we learn embeddings that are generally informative and this embeddings are shared globally across all tasks.

As we can see in Figure \ref{fig:model comparison} all our baseline models have embeddings that are shared across all tasks. With shared embeddings of transactions they obtain single user representation which is used to predict all attributes. However with this structures, models can't capture task specific features. In next subsection we will introduce embedding transformation layer to solve this problem. 

\subsection{Embedding Transformation Layer}
\label{section:embedding Transformation layer}

While the shared embedding layer captures features that are globally informative in all tasks, the embedding transformation layer is responsible for capturing task-specific features that are not shared across tasks. The embedding transformation layer changes the vector space of shared embeddings into a vector space of task-specific features. 
Embedding transformation is operated as follows:

\begin{equation}
\mathrm{V}^{m} = \mathrm{E}\mathrm{T}^{m}
\end{equation}
where $m$ $\in$ [$1$,...,$\mathrm{M}$] is an index of a task. The matrix $\mathrm{T}^{m} \in \mathcal{R}^{\mathrm{k} \times \mathrm{d}}$ is composed of trainable parameters. $\mathrm{k}$-dimensional shared embeddings are converted to task-specific embeddings with size $\mathrm{d}$. As $\mathrm{T}^m$ is  optimized only for the $m$-th task,  each converted embedding would form a task-oriented vector space. 

Linear transformation of this layer can also be viewed as a mapping function. Linear transformation maps globally shared embeddings to task-specific embeddings. By obtaining embeddings in a task-specific vector space based on shared features, our model can learn both general and more task-specific information.

\begin{figure}[t]
    \begin{center}
    \includegraphics[width=8.5cm,height=5.5cm]{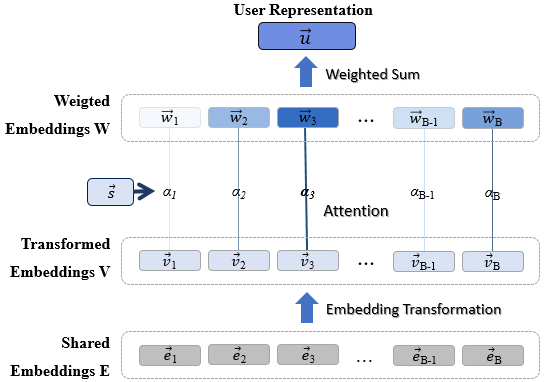}
    \caption{Detailed illustration of ETNA.}
    \label{fig:model detail}
    \end{center}
\end{figure}

\subsection{Task-Specific Attention Layer}
\label{section:task-specific attention layer}

Note that some transactions are more strongly associated with certain demographic attributes than others. To model this, we adopted task-specific attention mechanism.
The transformed embeddings [$\mathrm{V}^{1}$,...,$\mathrm{V}^{\mathrm{M}}$] are fed into a task-specific attention layer.
We obtain attention weights for each task by:

\begin{equation}
    \label{eq:attention score}
  z^{m} = f({\mathrm{V}^{m} \vec{s}^{m}} + b^{m})
\end{equation}

\begin{equation}
    \label{eq:attention weight}
  \alpha^{m} = softmax(z^{m}),
\end{equation}
where $f$ is a non-linear activation. 
$\vec{s}^{m} \in \mathcal{R}^{\mathrm{d}}$ and $b^{m} \in \mathcal{R}$ are trainable parameters that are not shared across tasks. In Equation \ref{eq:attention score}, the addition symbol $+$ denotes a broadcasting operation. The attention score $z^m$ is computed based on the similarity between the transformed embeddings $\mathrm{V}^m$ and $\vec{s}^m$. Distribution of attention weights $\alpha^{m} \in \mathcal{R}^{\mathrm{d}}$ describes the importance that our model assigned to each transaction.

 Equation (\ref{eq:attention score}), Equation (\ref{eq:attention weight}) are typically used to calculate attention weights with dot product. 
Our dataset contains the same repeated transactions, so calculating the attention weight using Equation (\ref{eq:attention score}) and Equation (\ref{eq:attention weight}) is not effective. When we divide each un-normalized value by the sum of all repeated transaction weight values, the distribution of attention weights would be flattened, reducing the variation in attention scores. Thus, to make attention mechanism more discriminating, we convert the list of transactions into an unique set.

Finally, we calculate the weighted sum over all transformed embeddings as follows:

\begin{equation}
  \vec{u}^{m} = \sum_{i=1}^{\mathrm{B}} \alpha_{i}^{m} \vec{v}_{i}^{m}
\end{equation}
where $\vec{u}^{m} \in \mathcal{R}^{\mathrm{d}}$ denotes the user representation for the $m$-th task, which means that each user is represented differently for each task. Figure \ref{fig:model detail} illustrates how ETNA obtains task-specific user representations.

The task-specific attention layer not only improves performance by utilizing useful data signals, but also makes our model more interpretable. We can check the attention weight distribution $\alpha=[\alpha^1,...,\alpha^{\mathrm{M}}]$ to see the signals that the model has focused more on. Further analysis on attention mechanism is presented in Section \ref{section:analysis}.

\subsection{Prediction Layer}
\label{section:prediction layer}

Finally, we obtain the predicted probability for the $m$-th demographic attribute of a given user by:

\begin{equation}
  p(y^{m}|\mathbf{x}) = softmax(\mathrm{O}^{m}\vec{u}^{m}),
\end{equation}
where $\mathrm{O}^{m} \in \mathcal{R}^{\mathrm{C}^{m} \times \mathrm{d}}$ is a trainable parameter. The parameter is responsible for converting the user representations for each task $\vec{u}^1,...,\vec{u}^{\mathrm{M}}$ into predictions through linear transformation.

The goal of demographic  prediction is to infer all demographic attributes of users from their transaction histories. Now, we recover the subscript $n$. For N users, we minimize the sum of the negative log-likelihoods defined as:

\begin{equation}
\label{eq:loss}
  - \sum_{n=1}^{\mathrm{N}} \sum_{m=1}^{\mathrm{M}} log p(y_{n}^{m}|\mathbf{x}_n) + \lambda||\Theta||,
\end{equation}
where $\Theta$ denotes all trainable parameters and $\lambda$ is the weight decay coefficient which is one of the hyperparameters. 
As mentioned in Section \ref{section:embedding Transformation layer} and Section \ref{section:task-specific attention layer}, total loss in Equation \ref{eq:loss} is calculated by summing the losses of each task; thus, the gradients only flow within the scope of corresponding tasks.

\section{Experiments}
\label{section:exp}
\subsection{Datasets}
\label{section:datasets}

We use the transaction dataset collected by a Korean multi vendor loyalty program provider. When customers use services or purchase items at a company contracted with this provider, a certain portion of the price is saved as points in the database. Customers can use these points like cash at all the stores that are enrolled in the program.
The dataset consists of purchasing histories of  56,028 users and contains the gender, age, and marital status (demographic attributes) of all the users. A total of 494 companies participate in the program. In our dataset, transactions are recorded as a [user ID, company ID, purchased amount] triplet. Although company names are hidden due to privacy issues, all industrial categories of companies are provided so that we can further analyze the behavior of our model. To the best of our knowledge, our dataset is the first public dataset containing both transaction meta-data and demographic information. We made the dataset publicly available\footnote{https://github.com/dmis-lab/demographic-prediction}. The statistics of our dataset are summarized in Table \ref{table:dataset}. 

As we described in Section \ref{section: task description}, we conducted experiments in two different problem settings. In the partial label prediction problem, our goal is to predict the unknown attributes of users while our model is trained with the observed attributes. All the users in our dataset have all demographic attributes. For the partial label prediction problem setting, we randomly set certain attributes as observed ($\mathbf{Y}^{\texttt{obs}}$) and used them in the training phase. We used an observed ratio of 10\% to 90\% with a step length of 10\%. When the observed ratio is 50\%, each attribute of a user has a 50\% chance to be observed and used in training. The remaining unknown attributes are used in the evaluation. To minimize noise due to randomness, we create 10 different splits for each observed ratio. We averaged the results of 10 datasets and report them in this paper.

For experiments on new user prediction, we split our dataset into non overlapping sets. We choose 8:1:1 as training, validation and testing split ratio, which results 44,822 users for training and 5,603 users for validation and testing respectively.

\begin{table}[]
\centering
\begin{tabulary}{\linewidth}{c|C|C}
    \toprule \hline
    Attributes & Classes & Distributions \\
    \hline \hline
    \multirow{2}{*}{Gender} 
    & Female & 63.0\% \\
    
    & Male & 37.0\% \\
    \hline
    \multirow{4}{*}{Age} 
    & Young & 22.3\% \\
    
    & Adult & 54.1\% \\
    
    & Middle Age & 14.3\% \\
    
    & Old & 9.4\% \\
    \hline
    \multirow{2}{*}{Marital Status} 
    & Married & 19.9\% \\
    
    & Single & 80.1\% \\
    \hline \bottomrule
\end{tabulary}
\caption{A description of our dataset.}
\label{table:dataset}
\end{table}

\subsection{Evaluation Metrics}
\label{section:metrics}
We employ several metrics appropriate to evaluate our model. These metrics are widely used in demographic prediction tasks. 

\begin{itemize}
\item \textbf{Hamming Loss} (HL) is a metric used to calculate the number of times predictions are incorrectly classified given a set of labels, which is defined as:
\begin{equation}
  \mathrm{HL} = \frac{1}{\mathrm{N}^{'}} \sum_{n=1}^{\mathrm{N}} \left( \frac{|\ddot{\mathbf{y}}_{n}\Delta
  \hat{\mathbf{y}}_{n}|}{|\ddot{\mathbf{y}}_{n}|} * \mathds{1}_{|\ddot{\mathbf{y}}_{n}| \neq 0} \right)
\end{equation}
where $\Delta$ is the symmetric difference and $|\cdot|$ denotes the size of a set inside the operation. $\mathds{1}$ is an indicator function. $|\ddot{\mathbf{y}}_n|$ is the number of attribute labels to be predicted and  is always set to $\mathrm{M}$ in new user prediction.
However, $|\ddot{\mathbf{y}}_n|$ changes in partial label prediction where labels are randomly observed. When all attributes of a user are observed, $|\ddot{\mathbf{y}}_n|$ can equal to 0. Hence, we calculate the metric over only $\mathrm{N}^{'}$ whose $|\ddot{\mathbf{y}}_n|$ is not 0.

\item \textbf{F1 Score} or \textbf{F-measure} is a widely used measure as a complement for accuracy. F1 score is usually calculated as the harmonic mean of precision (P) and recall (R):

\begin{equation}
  \mathrm{F1} = \frac{2 * \mathrm{P} * \mathrm{R}}{\mathrm{P} + \mathrm{R}}
\end{equation}

The precision and recall are formulated differently depending on the following types of F1 score: micro, macro and weighted. In our experiment, we do not use micro F1 since the goal of demographic prediction is to predict all combinations of attributes, not individuals. The precision and the recall of macro F1 (macF1) and weighted F1 (wF1) are computed as follows:

\begin{equation}
  \mathrm{P} = \sum_{\mathbf{y} \in \mathbf{Y}} \left( \frac{\sum_{n}\mathds{1}_{\hat{\mathbf{y}}_{n} = \ddot{\mathbf{y}}_{n} \& \mathbf{y} = \ddot{\mathbf{y}}_{n}}}{\sum_{n} \mathds{1}_{\mathbf{y}=\hat{\mathbf{y}}_{n}}} * weight \right) 
\end{equation}

\begin{equation}
  \mathrm{R} = \sum_{\mathbf{y} \in \mathbf{Y}} \left( \frac{\sum_{n}\mathds{1}_{\hat{\mathbf{y}}_{n} = \ddot{\mathbf{y}}_{n} \& \mathbf{y} = \ddot{\mathbf{y}}_{n}}}{\sum_{n} \mathds{1}_{\mathbf{y}=\ddot{\mathbf{y}}_{n}}} * weight \right)
\end{equation}
where $\mathbf{Y}$ is a set of all label combinations to be predicted. When employing macro F1, $weight = \frac{1}{|\mathbf{Y}|}$; otherwise, $weight = \frac{\sum_{n}^{\mathrm{N}} \mathds{1}_{\mathbf{y} = \mathbf{y}_{n}}}{\mathrm{N}}$. The macro F1 is an appropriate measure when considering each class as important as others. One characteristic of the macro F1 is that it is highly influenced by the minor classes. On the other hand, the weighted F1 assigns a high weight to the large classes by multiplying the number of classes by the weight. We report both metrics to compare the models.

\end{itemize}

\begin{figure*}[t]
    \begin{center}
    \includegraphics[width=18cm,height=5.25cm]{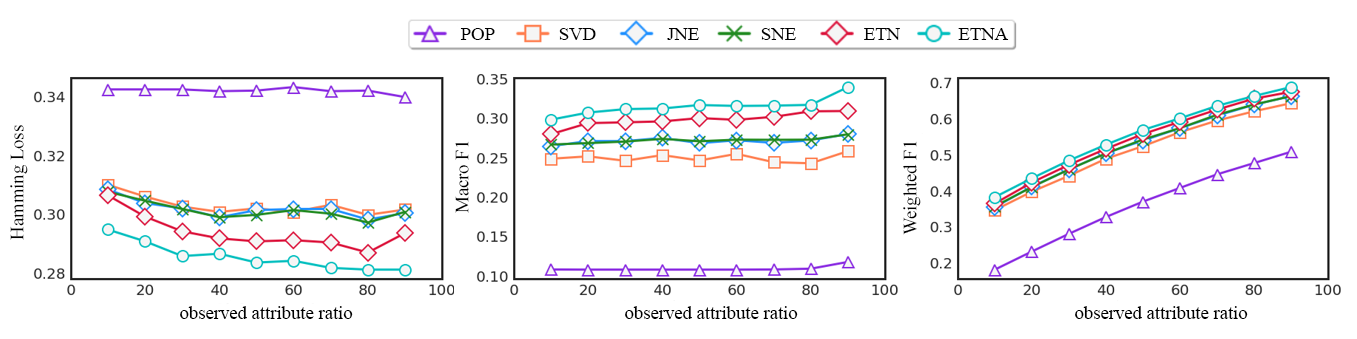}
    \caption{Comparison of different models in partial label prediction with varied observed attribute ratio from 10\% to 90\%.}
    \label{fig:result}
    \end{center}
\end{figure*}

\begin{table*}[h]
\centering
\footnotesize
\begin{center}
\begin{tabular}{c|ccccccc|ccccccc}
\toprule
& \multicolumn{7}{c|}{Partial Label (50\%)} & \multicolumn{7}{c}{New User} \\
Model & HL & macP & macR & macF1 & wP & wR & wF1 & HL & macP & macR & macF1 & wP & wR & wF1 \\
\hline\hline
 POP  & 0.342 & 0.082 & 0.159 & 0.108 & 0.289 & 0.514 & 0.370
 & 0.344 & 0.018 & 0.062 & 0.028 & 0.086 & 0.294 & 0.134 \\
 SVD    & 0.302 & 0.274 & 0.224 & 0.247 & 0.490 & 0.562 & 0.524 
 & 0.309 & 0.125 & 0.112 & 0.118 & 0.275 & 0.344 & 0.306  \\
 JNE    & 0.301 & 0.323 & 0.230 & 0.269 & 0.518  & 0.563 & 0.539 
 & 0.307 & 0.188 & 0.110  & 0.139 & 0.321 & 0.348 & 0.334  \\
 SNE    & 0.300 & 0.324 & 0.233 & 0.271 & 0.521 & 0.563 & 0.542 
 & 0.306  & 0.157 & 0.121 & 0.137  & 0.295 & 0.351 & 0.321  \\
 ETN & 0.291 & 0.350 & 0.263 & 0.300 & 0.540 & 0.576 & 0.557 
 & 0.295 & 0.193 & 0.144 & 0.165 & 0.310 & 0.368 & 0.336 \\
 ETNA    & \textbf{0.284} & \textbf{0.360} & \textbf{0.283} & \textbf{0.317} & \textbf{0.554} & \textbf{0.584} & \textbf{0.569} 
 & \textbf{0.286} & \textbf{0.216} & \textbf{0.156} & \textbf{0.182} & \textbf{0.339} & \textbf{0.382} & \textbf{0.360} \\

\bottomrule
\end{tabular}
\end{center}
\caption{The experimental results in partial label prediction (50\%) and new user prediction.}
\label{table:result}
\end{table*}

\subsection{Baseline Models}
\label{section:baselines}
To verify the effectiveness of our work, we compare our models with four baseline models.
The description of our models and the baseline models are listed below.

\begin{itemize}
\item \textbf{POP} : POP is a simple model that always predicts given users' attributes as the majority classes. POP is used in \cite{wang2016your} as a baseline model that ignores users' characteristics.
\item \textbf{SVD} : A singular value decomposition (SVD) is first conducted over user-item-matrix to obtain low dimensional representations of users. Logistic models are trained for each demographic attribute separately. This approach is widely used in demographic prediction tasks \cite{hu2007demographic, zhong2015you}.
\item \textbf{JNE} (Joint Neural Embedding) \cite{wang2016your} : In JNE each company has its own latent vector. JNE maps all transactions in users' histories into latent vectors. Average pooling is conducted to these vectors and then fed into a linear prediction layer for each task. The total loss is the sum of each loss as in Equation \ref{eq:loss}.
\item \textbf{SNE} (Structured Neural Embedding) \cite{wang2016your} :  SNE has very similar structures with JNE. The only difference between JNE and SNE is that the loss function of SNE is designed to be effective in multi-task learning problem. In SNE, the loss is computed via a log-bilinear model with structured predictions and labels (combination of attribute is considered). The model architectures of JNE and SNE are illustrated in Figure \ref{fig:model comparison}-(a),(b).
\item \textbf{ETN} (Embedding Transformation Network) : 
ETN is our proposed model with only embedding transformation layer is included. 
The output of the the embedding transformation layer $\mathrm{V}$ is directly fed into the prediction layer $\mathrm{O}$.
\item \textbf{ETNA} (Embedding Transformation Network with Attention):
 ETNA is our proposed model with all components included. Difference between ETN and ETNA is whether task-specific attention layer is included or not.
\end{itemize}

\begin{table*}[h]
\centering
\begin{center}

\begin{tabular}{c|ccc|ccc|ccc}
\toprule      & \multicolumn{3}{c|}{HL} & \multicolumn{3}{c|}{macF1} & \multicolumn{3}{c}{wF1} \\

ratio(\%) & JNE  & SEP  & ETN  & JNE  & SEP  & ETN  & JNE  & SEP  & ETN  \\
\hline
10    & 0.308 & 0.308 &  \textbf{0.306} & 0.264 & 0.262 & \textbf{0.280} & 0.355 & 0.357 &  \textbf{0.365}      \\
50    & 0.301 & 0.301 &  \textbf{0.291} & 0.268 & 0.269 & \textbf{0.300} & 0.539 & 0.540 & \textbf{0.557}   \\
90    & 0.300 & 0.301 &  \textbf{0.293} & 0.280 & 0.284 & \textbf{0.309} & 0.662 & 0.661 & \textbf{0.674} \\
\hline
\bottomrule
\end{tabular}
\end{center}
\caption{Comparison of different sharing structures in partial label prediction.}
\label{table:sharing result}
\end{table*}

\subsection{Experimental Settings}
\label{section:exp settings}

We implemented all the baseline models as described in \cite{wang2016your}.
We  have searched hyperparameters which performs best for our models  and  the  baseline  models  respectively.
We set the size of embedding $\mathrm{d}$ to 100 and the weight decay coefficient to $1e^{-5}$.
We use the Adam optimizer \cite{kingma2014adam}, with a mini-batch size of 64 and a learning rate of $1e^{-3}$. 
We apply early stopping for every epoch based on the weighted F1. The training process of partial label prediction (50\%) takes roughly 30 minutes on a single Titan X (Pascal) GPU and requires approximately 5GB of memory. The source code for the experiments is implemented using PyTorch\footnote{The source code is available at \textit{https://github.com/dmis-lab/demographic-prediction}}. 

\begin{figure*}[t]
    \begin{center}
    \includegraphics[width= 14.2 cm,height=4.7cm]{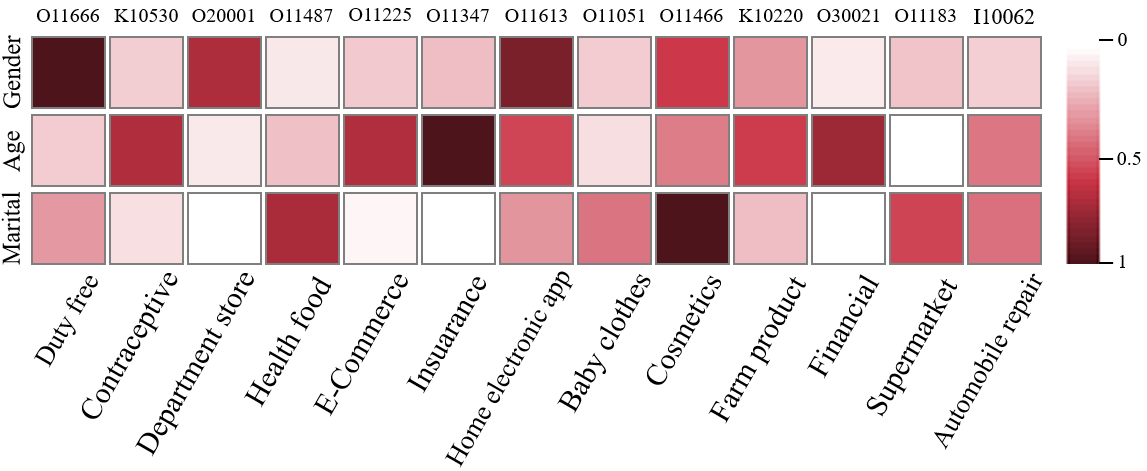}
    \caption{Comparison of attention weights calculated by ETNA}
    \label{fig:attention_vis1}
    \end{center}
\end{figure*}

\subsection{Results}
\label{section:results}

Table \ref{table:result} shows the experimental results on partial label prediction problem whose observed ratio of each attribute is 50\% and the new user prediction problem. We also included results on the partial label prediction problem for different observed ratios in Figure \ref{fig:result}.
Our baselines take different approaches to obtain user representation. Based on our experiment results, we have three findings about how different approaches behave.

(1) As JNE and SNE learn task oriented representations, both models perform better than SVD. However, SNE does not significantly outperform JNE compared to the results reported in \cite{wang2016your}. In our experiment on new user prediction, JNE even outperforms SNE. As we mentioned before, JNE and SNE are very similar especially in obtaining user representations and predicting attributes from obtained user representations. Changing the objective function had a minimal effect on our experiment. 

(2) In all the experiments, our models outperform all the baselines. As we emphasized throughout this entire paper, the ability to learn task-specific features is also important in multi-task learning problems. Although both JNE and ETN predict each attributes using separate prediction parameters, JNE's capability is limited because it uses single user representation. On the other hand, ETN utilizes task-specific user representations obtained from embeddings which are optimized only for individual tasks. Attention mechanism in our model improves the performance of our model even further by focusing more on important signals in a task specific manner. ETNA achieves the best accuracy in all our experimental settings. 

(3) To demonstrate the impact of embedding transformation in the demographic prediction task, we included the experimental results of models with different sharing structures in Table \ref{table:sharing result}. SEP is a model with the loss function defined as Equation \ref{eq:loss} but with a separate embedding for each task, which indicates that SEP does not share any feature across tasks. As our experimental results demonstrate, JNE and SEP obtain very similar performance. Both models have their own strengths (shared information,  capturing task specific features) but without the ability to balance these two properties they cannot achieve better accuracy. Our ETN model  outperforms both models in all experiments. The results show that obtaining task-specific user representations with embedding transformation is simple but effective.

We also conducted experiment on Beiren dataset used in \cite{wang2016your}. However, we couldn't reproduce the result reported in \cite{wang2016your}. We got slightly but not significantly higher score than JNE and SNE. Based on our results and analysis, item level transaction data in Beiren doesn't have enough information to predict demographic attributes. Experiment results on Beiren dataset are provided in Appendix.

\subsection{Visualization of Task-Specific Attention}
\label{section:analysis}

To further analyze the impact of attention mechanism in our model, we provide visualization of weights calculated by attention mechanism. We obtained attention scores for each company. As attention mechanism is independent for each task, each company is assigned different weight for each task. We picked examples that provide insights for customer behaviors from 20 companies with highest attention scores in each task. Based on the attention weights from our model, we draw heatmap in figure \ref{fig:attention_vis1} . 

First thing to notice in this example is that duty-free brand obtained a highest attention in gender prediction task. As cosmetics and perfumes are most popular products in duty free stores, we might say females purchase ore actively as they are generally more interested in those items. The fact that same brand got relatively lower score in age prediction task also accords with our intuition.

In age prediction, ETNA gave high attention scores to insuarance and automobile related services. Intuitively, older people have more chance to have their own cars and also are more interested in protecting themselves from possible risks. 

Interesting point here is that some brands obtained high attention scores in multiple tasks and some brands are focused only in specific task. For example, home electronic appliance company was assigned high scores in all tasks. However, attention scores of health food and contraceptive are very different for each task.

Some results which are not expected by us before the analysis. Although our common sense accords with high score for cosmetic brand in gender prediction task, we didn't expect that the cosmetic brand would get a high score in marital status prediction task. By analyzing unexpected facts, we might get a deeper understanding about customer behaviors.

It is noteworthy that interpretations of why our model attended on some companies can vary. There is no special way to generate an unique explanation. Our goal is to provide business managers a tool to understand customer behaviors and establish strategy based on data-driven way. 

\section{Conclusion}
\label{section:conclustion}

In this paper, we studied the demographic prediction task in retail business scenario as a multi-task learning problem. We proposed an Embedding Transformation with Attention (ETNA) model which transforms shared embeddings into task-specific embedding and detects more important signals with attention mechanism. We demonstrated ETNA's improved accuracy over existing state-of-the-art demographic prediction models.  We analyzed attention weights to understand customer behaviors in a data-driven way. We also released our dataset collected by a multi-vendor loyalty service provider. 

\section*{Acknowledgement}
This work was supported by the National Research Foundation of Korea(NRF-2017R1A2A1A17069645, NRF-2017M3C4A7065887).


\bibliographystyle{siam}
\bibliography{Reference}
\end{document}